\documentclass[conference]{IEEEtran}
\ifCLASSINFOpdf
\else
\fi
\usepackage[cmex10]{amsmath}
\usepackage{framed}
\usepackage{graphicx}
\usepackage{caption}
\usepackage{subcaption}
\usepackage{subfig}

\usepackage[font=footnotesize]{caption}
\usepackage[]{algorithm2e}

\divide\pdfpxdimen by 300

\begin{document}
%
\title{Gibbs Sampling with \\ Low-Power Spiking Digital Neurons}

\author{\IEEEauthorblockN{Srinjoy Das$^{\dagger\,+}$, Bruno Umbria Pedroni$^\bot$, Paul Merolla$^\ddag$, John Arthur$^\ddag$, Andrew S. Cassidy$^\ddag$, Bryan L. Jackson$^\ddag$ \\ Dharmendra Modha$^\ddag$, Gert Cauwenberghs$^{\bot\,+}$, Ken Kreutz-Delgado$^{\dagger\,+}$}

\medskip
Email: \{s2das, bpedroni, kreutz, gert\}@ucsd.edu, \{pameroll, arthurjo, andrewca, bryanlj, dmodha\}@us.ibm.com
\IEEEauthorblockA{$^\dagger$ ECE, $^\bot$BioEng. \& $^+$Inst. for Neural Computation, UC San Diego, La Jolla, CA 92093\\
$^\ddag$IBM Research Almaden, San Jose, CA 95120\\
}
}


%


\maketitle

\begin{abstract}
Restricted Boltzmann Machines and Deep Belief Networks have been successfully used in a wide variety of applications including image classification and speech recognition. Inference and learning in these algorithms uses a Markov Chain Monte Carlo procedure called Gibbs sampling. A sigmoidal function forms the kernel of this sampler which can be realized from the firing statistics of noisy integrate-and-fire neurons on a neuromorphic VLSI substrate. This paper demonstrates such an implementation on an array of digital spiking neurons with stochastic leak and threshold properties for inference tasks and presents some key performance metrics for such a hardware-based sampler in both the generative and discriminative contexts. \\  
\end{abstract}


%
\IEEEpeerreviewmaketitle
\section{INTRODUCTION AND BACKGROUND}

Restricted Boltzmann Machines (RBMs)  and Deep Belief Networks (DBNs) (Fig. \ref{figure_RBM_DBN}) are stochastic neural networks that have been used for a wide variety of generative and discriminative tasks like image classification, sequence completion, motion synthesis and speech recognition. An RBM is a stochastic neural network consisting of two symmetrically interconnected layers composed of neuron-like units --- a set of visible units $v$ and a set of hidden units $h$. For an RBM there are no interconnections within a layer. Both inference and learning in this model use a Markov Chain Monte Carlo (MCMC) procedure called Gibbs Sampling \cite{haykin08neural} where each neuron is sampled based on its total input from other connected neurons with a sigmoidal activation function. DBNs consist of one visible layer and multiple layers of hidden units. Learning in a DBN can be done in a layer-by-layer manner on each RBM with this Gibbs sampling procedure \cite{hinton2006fast}. Following this approach the values of the neurons in each layer of a DBN can be inferred by using the same stochastic sampling procedure.

\medskip
\graphicspath{{/Users/rumpagiri/Documents/RESEARCH_2014/docs_papers/digital_gibbs/}}
\DeclareGraphicsExtensions{.pdf}


Neuromorphic computing is an area of Very Large Scale Integrated Circuit (VLSI) design inspired by the architecture and function of the brain. Such systems which have been realized with both analog \cite{indiveri2011neuromorphic} and digital \cite{merolla2014million} circuit elements consist of massively parallel arrays of interconnected spiking neurons modeled on the basis of neurons and synapses present in biological neural substrates. In contrast to the traditional von Neumann computing paradigm, memory and computation are tightly coupled in this architecture. The principal benefits are extremely energy efficient computation by spiking neurons in a highly concurrent fashion.

{\begin{figure}[!t]
  \centering
  \includegraphics[width=2.5in, height=1.5in]{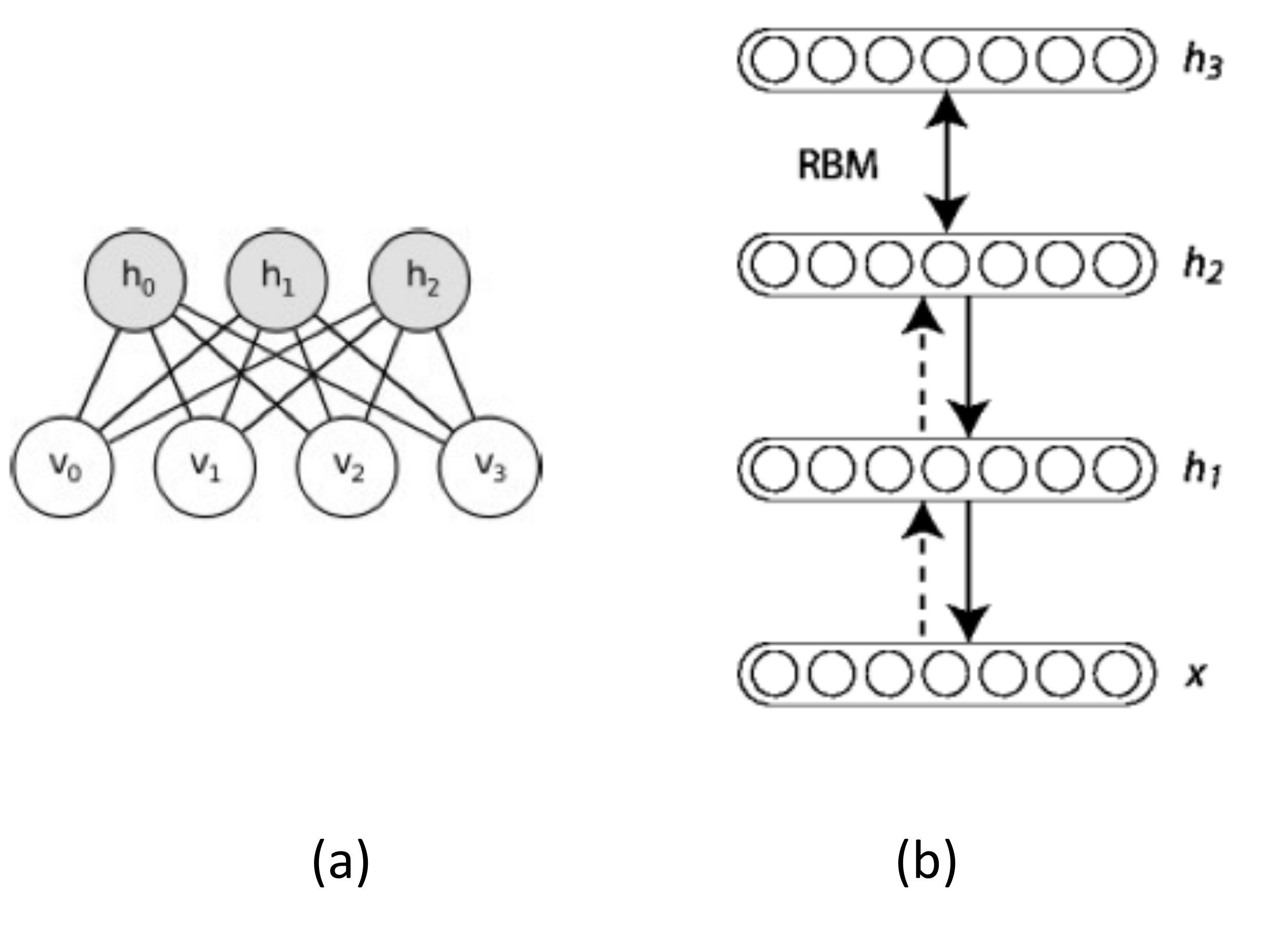}
  \caption{a) Restricted Boltzmann Machine with 4 visible and 3 hidden units. b) Deep Belief Network with 3 hidden layers} 
  \label{figure_RBM_DBN}
\end{figure}}

The majority of RBMs and DBNs described in the literature currently operate on standard platforms like high performance CPU (Central Processing Unit) or GPU (Graphical Processing Unit) and are deployable on cloud and related infrastructures. However, for ultra low-power, realtime realizations of these algorithms, hardware substrates provided by neuromorphic VLSI are naturally amenable to the use of sampling methods for probabilistic computation in the context of high-dimensional real world data. In this paper we propose an MCMC sampling scheme for RBMs and DBNs using the stochastic  leak and threshold properties of digital spiking neurons on a neuromorphic VLSI substrate. Such a framework has significant potential for enabling applications which will benefit from realtime, energy-efficient realizations, such as the Internet of Things and Brain Computer Interfaces.

\section{INFERENCE ON SPIKING SUBSTRATES}

The RBM captures a probabilistic generative model of the input data based on the Boltzmann distribution as below \cite{haykin08neural}:
\begin{align}
\begin{cases}
\bf p(v, h) = \frac {exp(-E(v, h))}{\Sigma _{v, h} exp(-E(v,h))}\\
\bf where \ E(v,h) = -v^{T}Wh - b_{v}^{T}v - b_{h}^{T}h
\label{eq:BD}
\end{cases}
\end{align}

Here $p$ denotes the Boltzmann probability distribution and $E$ is an energy function of $v$ and $h$ where $v$ denotes the state of the visible units which are driven by the input data and $h$ represents the state of the hidden units. $W$ represents the weight between $v$ and $h$, and $b_v$, $b_h$ represent the biases of $v$ and $h$ respectively. A necessary and sufficient condition for sampling from the Boltzmann distribution given by the above equation is to sample each neuron with a sigmoid probability law as a function of the activities of all other connected neurons \cite{rojas1996neural}. This is as given below :
\begin{align}
\bf P(x_{i} = 1 | x_{j}, j \neq i) = \frac {1} {1 + \ e^{-(\Sigma_{j}w_{ij}x_{j} + b_{i})}  }
\label{eq:sigmoid}
\end{align}

This rule forms the kernel of the Gibbs sampling MCMC procedure for an RBM. Here $w_{ij}$ is the weight between neurons $x_{i}$, $x_{j}$ and $b_{i}$  denotes the bias of neuron $x_{i}$. This equivalence between probability laws at the unit and ensemble levels is exactly realizable for substrates where explicit synchronization is provided and all samples from the underlying primitives representing neurons are collected in discrete steps. This allows the RBM to do alternating parallel sampling to generate statistics for the MCMC inference process.

\medskip
The TrueNorth neurosynaptic processor \cite{merolla2014million} is an array of 4096 cores. Each core consists of a crossbar with 256 axons and 256 neurons. Communication between neurons occurs only with all-or-none spikes realized with digital circuitry. In the interconnection network, the spikes are generated with asynchronous digital logic, however all spikes are explicitly aligned at a single clock edge for use in the next step of computation. On this substrate, the stochastic and dynamical properties of the neurons coupled with these discrete synchronization steps can be used to generate the sigmoid probability law used for sampling from the Boltzmann distribution of interest. A Gibbs sampler for inference in RBMs and DBNs can thus be constructed on such a substrate for the conditions outlined in Eqns. (\ref{eq:BD}), (\ref{eq:sigmoid}).

\section{DIGITAL GIBBS SAMPLER}

TrueNorth \cite{merolla2014million} is composed of digital integrate-and-fire neurons (I\&F) with both stochastic and deterministic leak and threshold properties. The dynamical equations for the membrane potential $V_{j}(t)$ for neuron $j$ at time $t$ in this case are shown below \cite{Cassidy13cognitivecomputing}:
\begin{align}
\begin{cases}
\bf V_{j}(t) = V_{j} (t-1) + \Sigma _{i=0}^{N-1} x_{i}(t)s_{i}\\
\bf V_{j}(t) = V_{j}(t) - \lambda_{j}\\
\bf If \ (V_{j}(t) \ge \alpha_{j}, \ SPIKE \ and \ set \ V_{j}(t) = R_{j}\\
\end{cases}
\label{eq:i_and_f}
\end{align}

Here $\lambda_{j}$ and $\alpha_{j}$ represent the leak and threshold values respectively of neuron $j$ which can be stochastic or deterministic, $x_{i}$ represents the input from $N$ other neurons, $s_{i}$ represents the synaptic weight between neurons $i$, $j$ and $R_{j}$ represents the reset value for neuron $j$.

Using similar I\&F neurons with stochastic leak and threshold an algorithm for realizing the sigmoidal sampling rule (Eqn. \ref{eq:sigmoid}) to perform MCMC sampling in RBMs  is given below:
\hspace{-1cm}
\begin{algorithm}
 \Repeat{$ \bf Tw \ steps$}{
 $ \bf V = V + leak*(B (0.5)),\  B \ is \ Bernoulli$ \\
$ \bf Vt\_rand = Vt + floor(U(0,2^{TM}-1))$ \\
$ \bf spiked(V \ge Vt\_rand) = 1 $\\
 }
\end{algorithm}
%
%

Each neuron in the RBM is mapped to one digital neuron in this sampling scheme. The algorithm uses 4 parameters: the number of discrete time steps $Tw$ for sampling, fixed threshold value $Vt$, the number of bits allowed for the stochastic threshold variation $TM$ (uniformly distributed discrete random variable) and the value of $leak$. For the sampled neuron, $Vt\_rand$ and $V$ denote its threshold and membrane potential respectively. After integration, the sampled value of a neuron is set to 1 if it spikes in $any$ of the allowed number of sampling intervals $Tw$. Note that $TM$ and $leak$ are both positive. This method uses the underlying substrate's dynamical (integration) and stochastic properties along with spike synchronization at fixed time steps to generate a sigmoid probability curve. Random threshold and leak values are realized with on-chip pseudo random number generators (PRNGs).

\section {WEIGHT AND BIAS SCALING}

\begin{figure}[!t]
  \centering
  \includegraphics[width=3.5in, height = 1.2in]{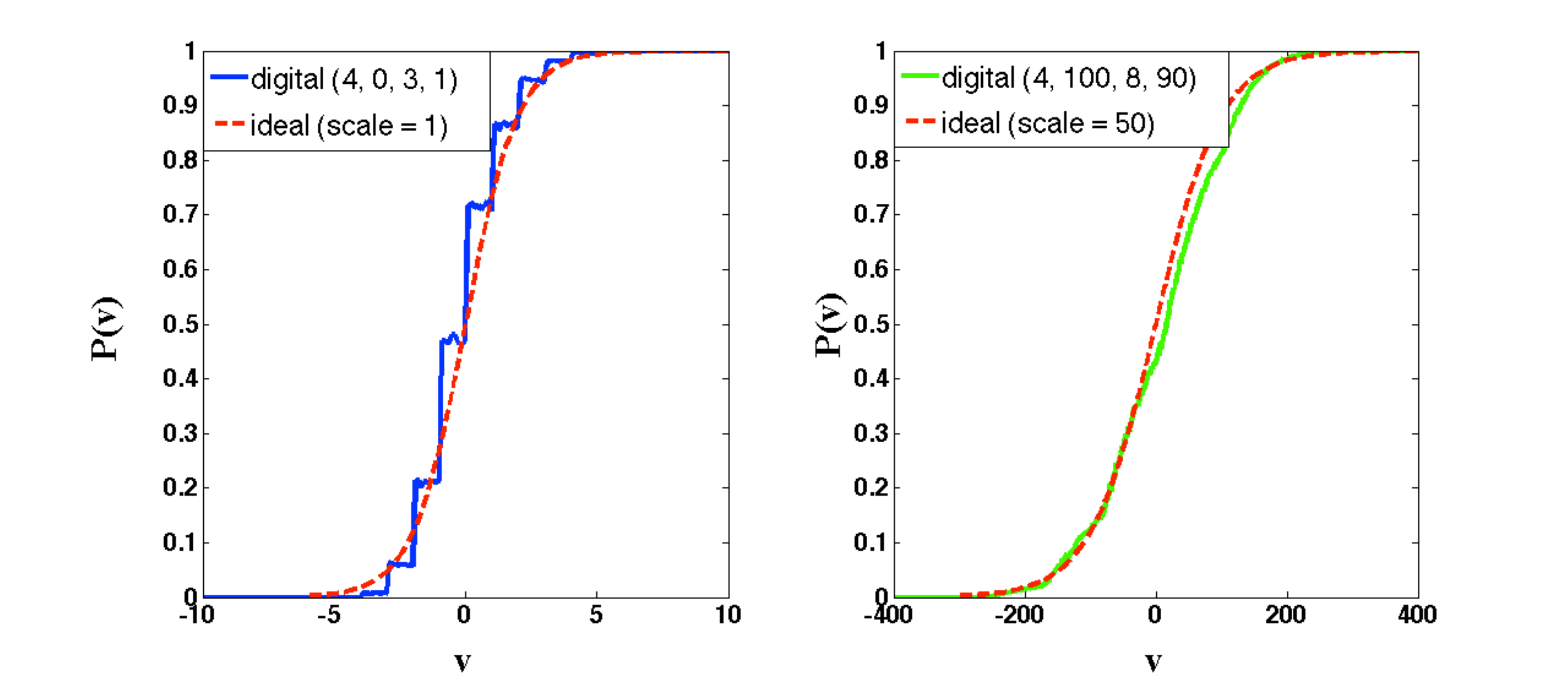}
  \caption{Noisy and smooth realizations of digital sigmoid using 1000 samples per v-value} \label{figure_sigmoid_smooth_noisy}
\end{figure}

On a substrate of digital neurons and synapses, weight and bias values have to be quantized in accordance with the finite precision available in hardware. To increase the dynamic range, a multiplicative  factor ($scale > 1$) is applied to the weights obtained after offline training of the RBM/DBN and before they are mapped to the hardware for applicable inference tasks (classification, pattern completion and others). Therefore the four sigmoid parameters ($Tw, Vt, TM$ and $leak$) should be chosen so that for any neuron $i$ there is a smooth mapping between the values of $v_i$ = $\Sigma_j w_{ij}x_j + b_{i}$ versus the activation probability $P(v_i)$ and a majority of the inputs are not mapped to the portion of the curve which saturates to 0 or 1. This is illustrated in 
Fig. (\ref{figure_sigmoid_smooth_noisy}),  where parameter values of (4,0,3,1) provide an insufficient range for smooth realization of the function. In contrast parameter values of (4,100,8,90) provide a larger dynamic range, where the original weight and bias values have been multiplied by a scale factor (in this case $scale$ = 50) for the mapping. This is equivalent to scaling the ideal sigmoid by this same factor:
\begin{align}
\bf P_{scaled}(v) = \frac {1} {1 + e^{-\frac{v}{scale}}  }
\label{eq:sigmoid_scale}
\end{align}

\graphicspath{{/Users/rumpagiri/Documents/RESEARCH_2014/docs_papers/digital_gibbs/}}
\DeclareGraphicsExtensions{.pdf}

\captionsetup{belowskip=-1pt,aboveskip=4pt}

 
\section {EFFECT OF STOCHASTIC PARAMETERS}

\begin{figure}
\centering
\begin{subfigure}{.5\textwidth}
  \centering
  \includegraphics[scale=0.18]{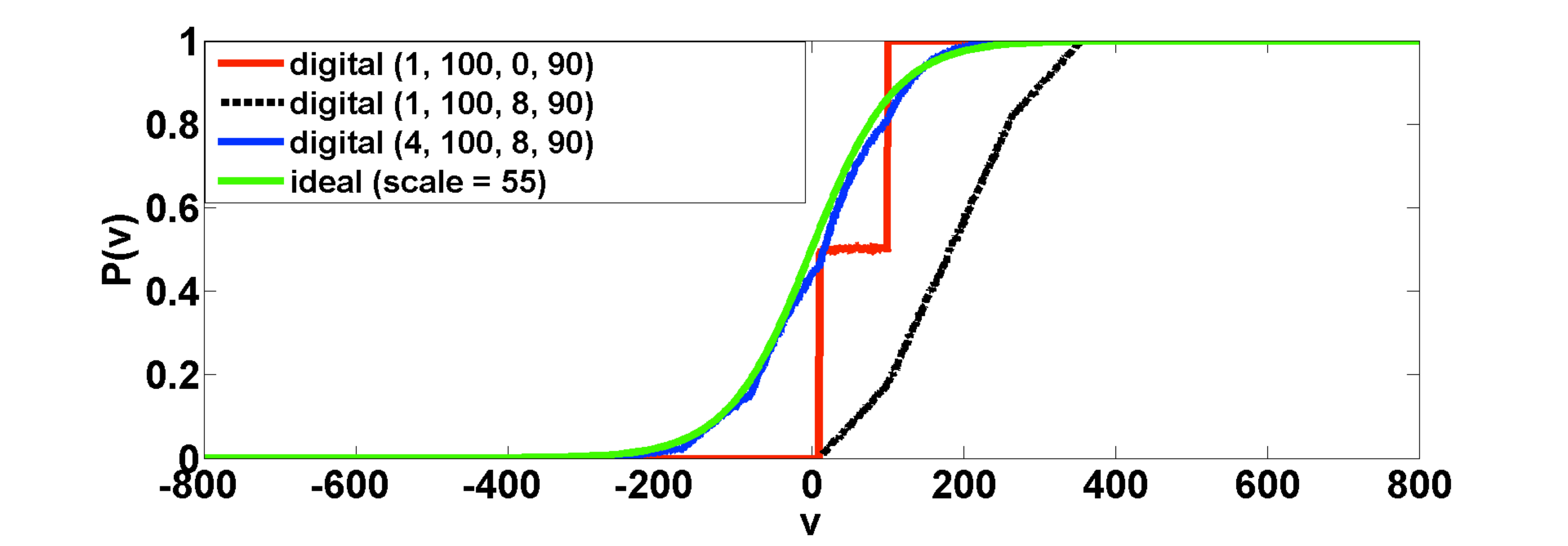}
  \caption{Effect of noisy threshold}
  \label{figure_sigmoid_noisy_threshold}
\end{subfigure}
\begin{subfigure}{.5\textwidth}
  \centering
    \includegraphics[scale=0.18]{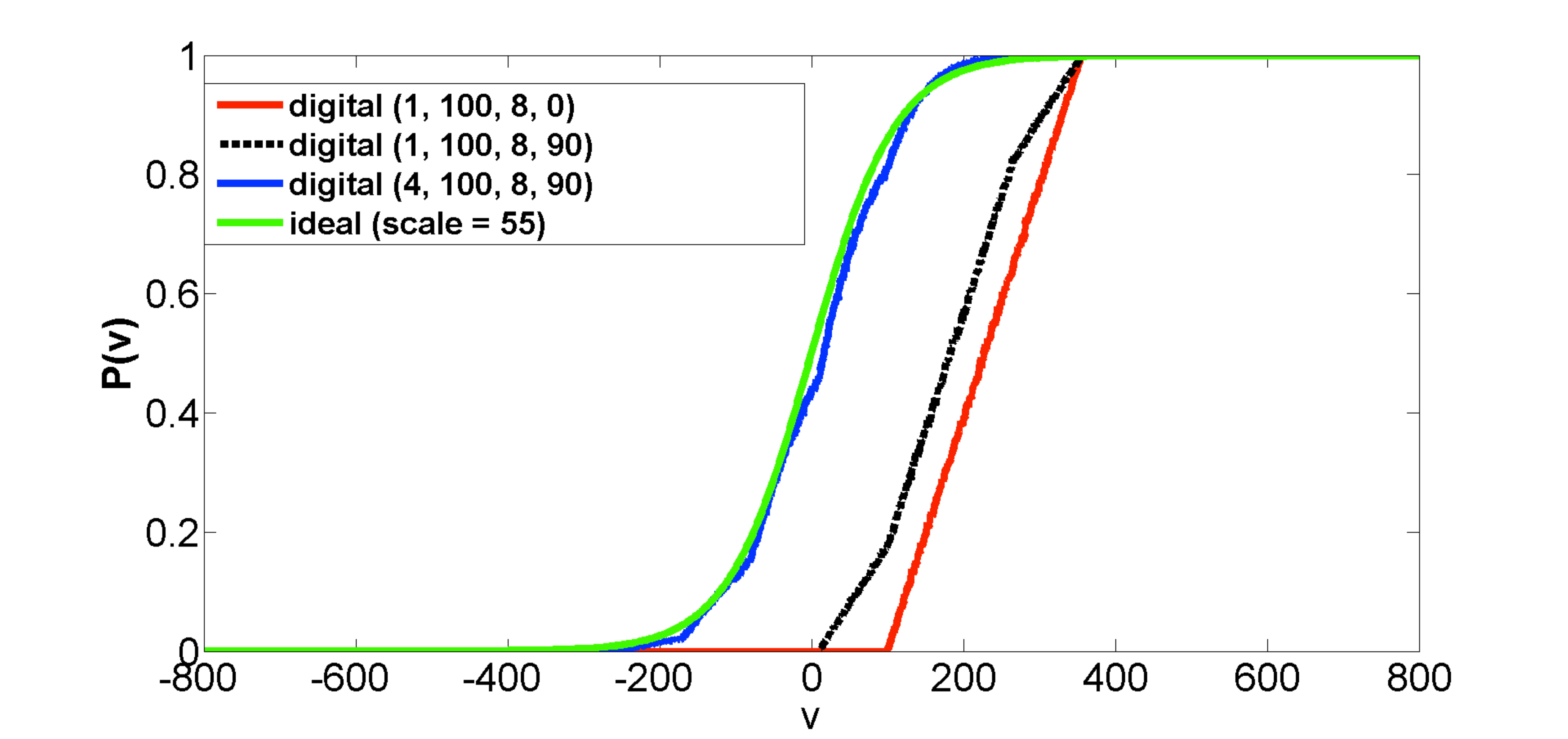}
  \caption{Effect of noisy leak}
  \label{figure_sigmoid_noisy_leak}
\end{subfigure}
\caption{Noisy threshold and leak for sigmoid realization}
\label{figure_sigmoid_noisy_threshold_leak}
\end{figure}

A heuristic motivation for the sigmoid realization using the stochastic leak and threshold over multiple sampling intervals is provided in Figs. (\ref{figure_sigmoid_noisy_threshold}, \ref{figure_sigmoid_noisy_leak}).  In the first case  Fig. (\ref{figure_sigmoid_noisy_threshold}) only the stochastic leak is initially applied to sample the inputs which causes the transfer curve to split into 3 regions:  input $<$ 10 where the neuron will never spike as the membrane potential is always below threshold, input $>$ 10 and $<$ 100 where the spiking probability is 0.5 (since the leak of 90 can occur with probability 0.5) and input $>$100 where the neuron is guaranteed to spike since the threshold is reached. As the stochastic threshold is applied, the curve becomes piecewise-linear which approximates to a sigmoid over multiple sampling intervals $Tw$. Similarly when only the stochastic threshold is initially applied (Fig. \ref{figure_sigmoid_noisy_leak}), the probability of activation is proportional to the difference between the input and the fixed threshold. As the stochastic leak is applied over multiple $Tw$ sampling intervals the curve also attains sigmoidal shape.

\graphicspath{{/Users/rumpagiri/Documents/RESEARCH_2014/docs_papers/digital_gibbs/}}
\DeclareGraphicsExtensions{.pdf}


\captionsetup{belowskip=-12pt,aboveskip=4pt}

\section {PERFORMANCE METRICS}

Different realizations of the sigmoidal approximation with varying complexity are possible for performing inference in RBMs/DBNs using Gibbs sampling. This can be studied in the context of both discriminative and generative tasks.

\graphicspath{{/Users/rumpagiri/Documents/RESEARCH_2014/docs_papers/digital_gibbs/}}
\DeclareGraphicsExtensions{.pdf}

\begin{figure}[ht]
\centering
\begin{minipage}[b]{0.45\linewidth}
\includegraphics[width=2.5cm]{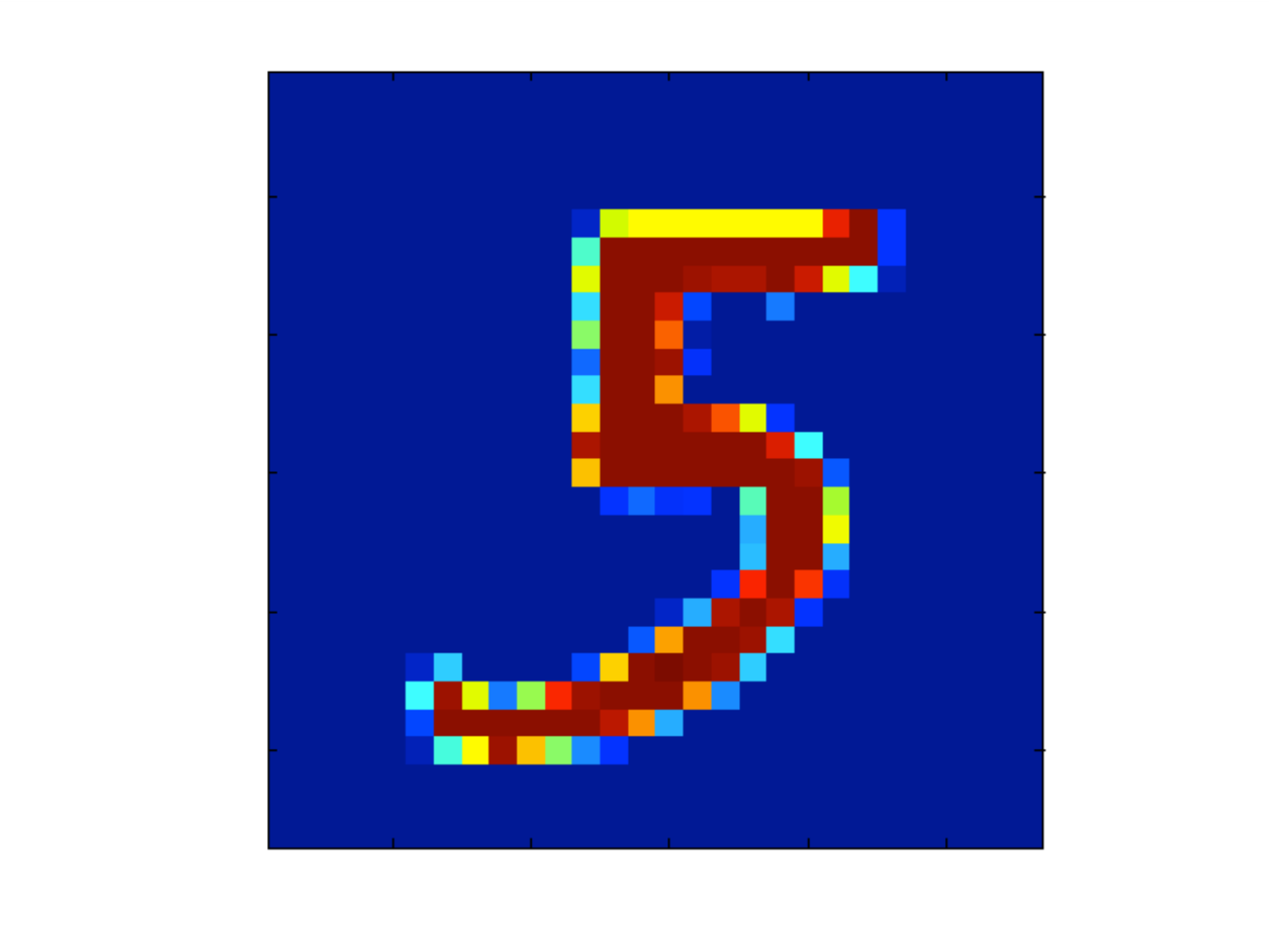}
\caption{Test data}
\label{figure_testdata}
\end{minipage}
\hfill
\begin{minipage}[b]{0.45\linewidth}
\includegraphics[width=2.5cm]{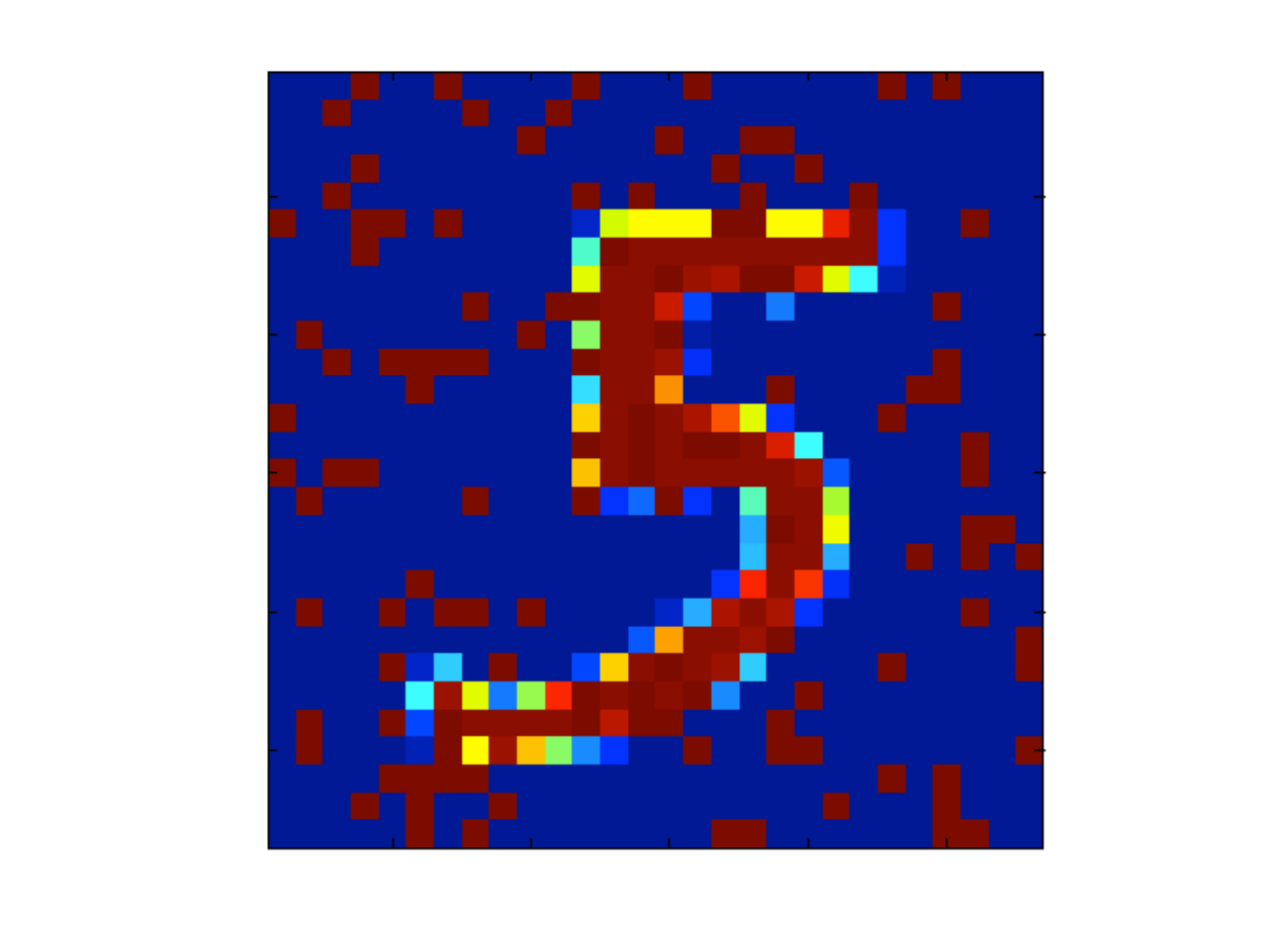}
\caption{Test data with salt noise}
\label{figure_testdata_salt}
\end{minipage}
\end{figure}

\subsection {CLASSIFICATION PERFORMANCE}

\begin{figure}[!t]
  \centering
  \includegraphics[width=3.5in, height = 1 in]{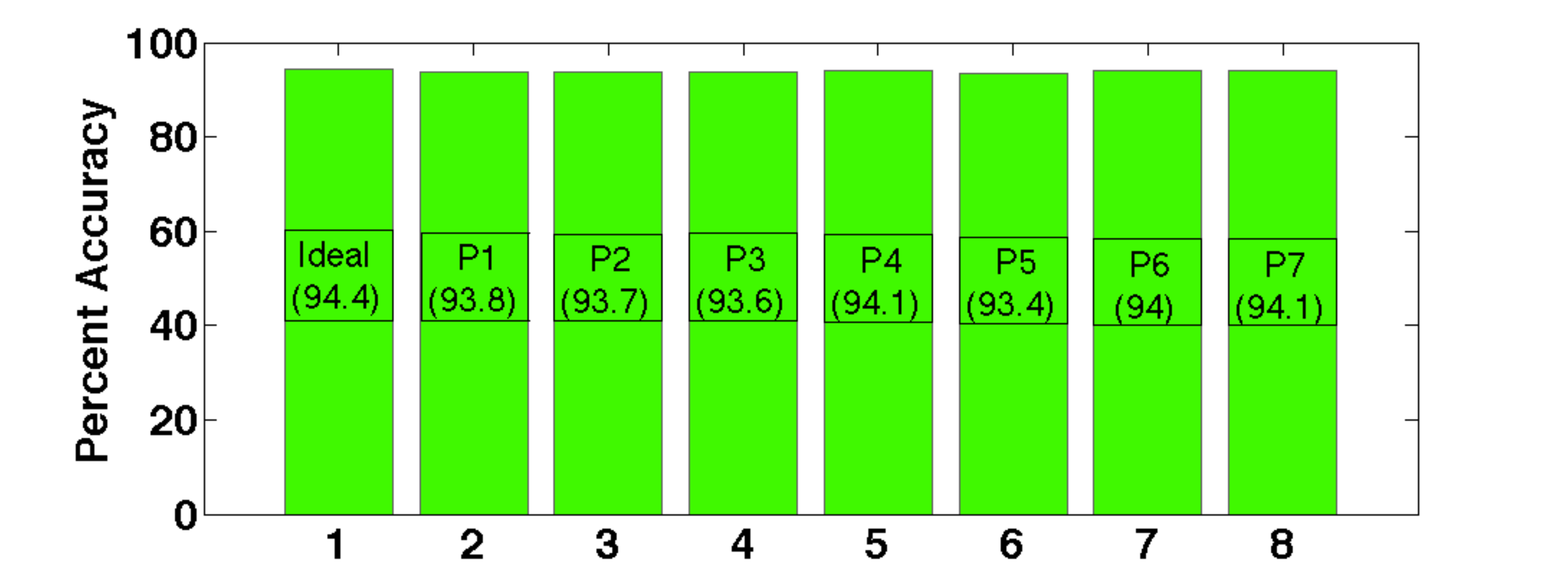}
  \caption{Classification accuracy for test data without noise} \label{figure_logistic_vs_digital_nonoise}
\end{figure}

For classification we study the performance of the digital sampler on the MNIST dataset which consists of 28x28 grayscale images of handwritten digits [0-9] (Fig. \ref{figure_testdata}) and their corresponding labels. The RBM with 784 visible, 500 hidden and 10 label neurons is trained offline on such a set of 5000 digits (\textbf {training data}). Classification performance is then tested for this RBM with the same number of visible, hidden and label units using the digital sampler with varying parameterizations of $ Tw, Vt, TM, leak$ (refer Table I) on a set of 1000 labeled digits (\textbf {test data}) which are similar to those seen by the RBM during training.  Appropriate scale factors are applied to the weights and biases as outlined in Section IV before the digital sigmoid is used for sampling. The classification results are shown in Fig. \ref{figure_logistic_vs_digital_nonoise}. It is noticeable that there is no significant difference in classification performance irrespective of the sampler complexity or the scale factor used for the digital realization.

\begin{table}[h!]
  \begin{center}
    \begin{tabular}{| l || c |r|}
    \hline
    Index &  (Tw, Vt, TM, leak) & scale\\
    \hline \hline
    P1 & (1, -130, 8, 0)  & 50\\
    P2 & (1, -80, 8, 102) & 50\\
    P3 & (1, -20, 8, 200) & 75\\
    P4 & (1, -100, 9, 300) & 120\\
    P5 & (16, 50, 9, 15) & 30\\
    P6 & (16, 100, 10, 30) & 50\\
    P7 & (16, 633, 8, 90) & 100\\
    \hline
    \end{tabular}
  \end{center}
  \caption{Digital neuron parameters}
\end{table}

\subsection {GENERALIZATION PERFORMANCE}

\begin{figure}[!t]
  \centering
  \includegraphics[width=3.5in]{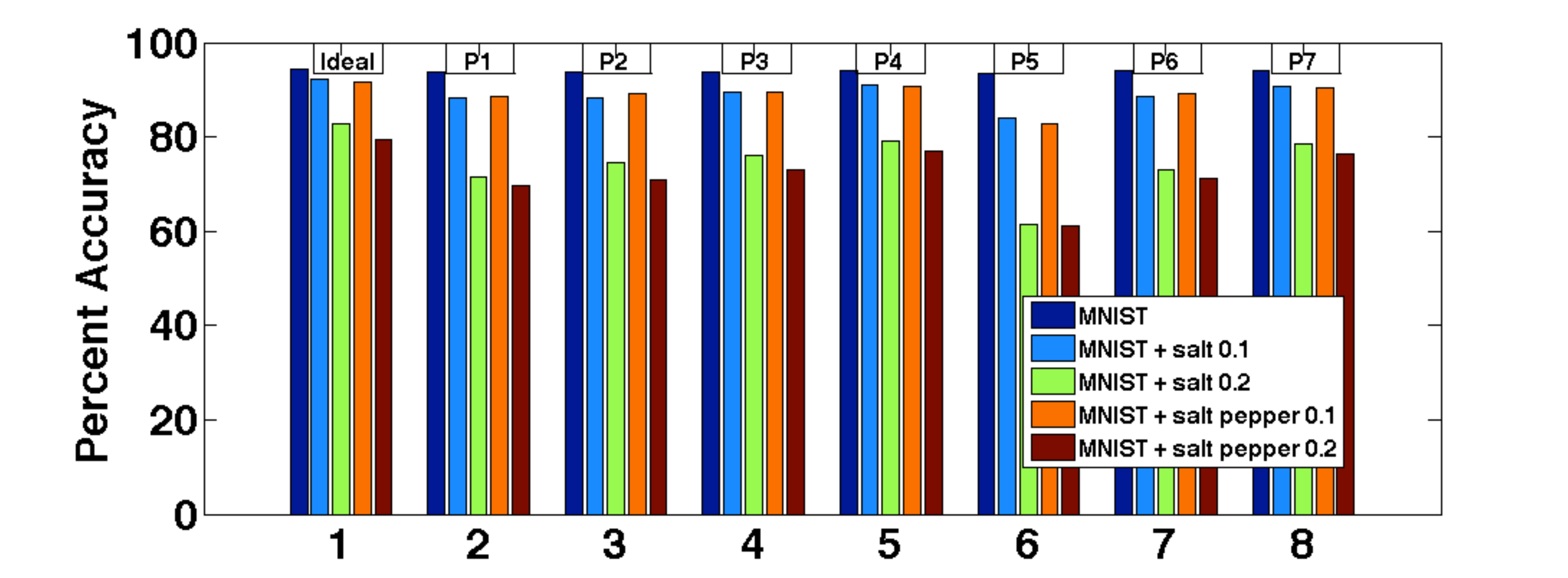}
  \caption{Classification accuracy for test data with noise} \label{figure_logistic_vs_digital_noise}
\end{figure}

The classification performance of different realizations of the digital sampler can also be studied with noisy versions of the test data. Two types of noise are introduced in the test dataset: \textbf {salt noise} which corrupts randomly chosen pixels to white and \textbf {salt and pepper noise} where randomly chosen pixels can be turned white or dark. In both cases the level of noise (pixel corruption) is controllable with a $noise\  factor$. This type of data (refer Fig. \ref{figure_testdata_salt}) is very different from those that were used for training the RBM/DBN, so this provides a measure of the generalization performance of the sampler. The classification results for an RBM with 500 hidden units are shown in Fig. \ref{figure_logistic_vs_digital_noise}. The sampler resolution (scale factor) has a significant effect on classification performance for noisy versions of \textbf{test data} as compared to the original \textbf{test data} itself. The samplers with the highest scale factor (P4, P7) have significantly fewer errors for the noisy data sets.

\subsection {GENERATIVE MODEL PERFORMANCE}

Inference performed on an RBM in hardware for generative tasks like pattern completion depends on the quality of the MCMC samples. That is, how closely these samples reflect the learned probability distribution. This can be characterized by the Kullback-Leibler (KL) divergence metric which is a similarity measure between any 2 probability distributions. MCMC sampling with the ideal and various digital realizations of the sampler can be used to generate the probability distributions and these can be compared versus the exact distribution given by Eq. (\ref {eq:BD}). Since it is difficult to calculate this distribution for high-dimensional inputs like MNIST (784 visible neurons), this is done for an RBM with 3 visible and 2 hidden neurons with pre-defined weights and biases. The results are shown in Table II. The KL-divergence increases for approximate realizations of the digital sampler ($leak$ = 0) indicating a lower quality of the sampled generative model in this case.

\begin{table}[h!]
  \begin{center}
    \begin{tabular}{| l || l | l | l | l |}
    \hline
    KL divergence (1e+05 Gibbs iterations) & Trial-1 & Trial-2 & Trial-3\\
    \hline \hline
    exact vs ideal sampler & 6.2e-05 & 6.06e-05 & 5.71e-05\\
    exact vs digital(1,-130,8,0) & 0.0218 & 0.1090 & 0.0259\\
    exact vs digital(1,-80,8,102) & 0.0091 & 0.0330 & 0.0083\\
    \hline
    \end{tabular}
  \end{center}
  \caption{KL divergence for various digital sampler realizations}
\end{table}

\section {TRUENORTH SIGMOID CHARACTERIZATION }

\begin{figure}[!t]
  \centering
\includegraphics[width=3.4in, height=2.4in]{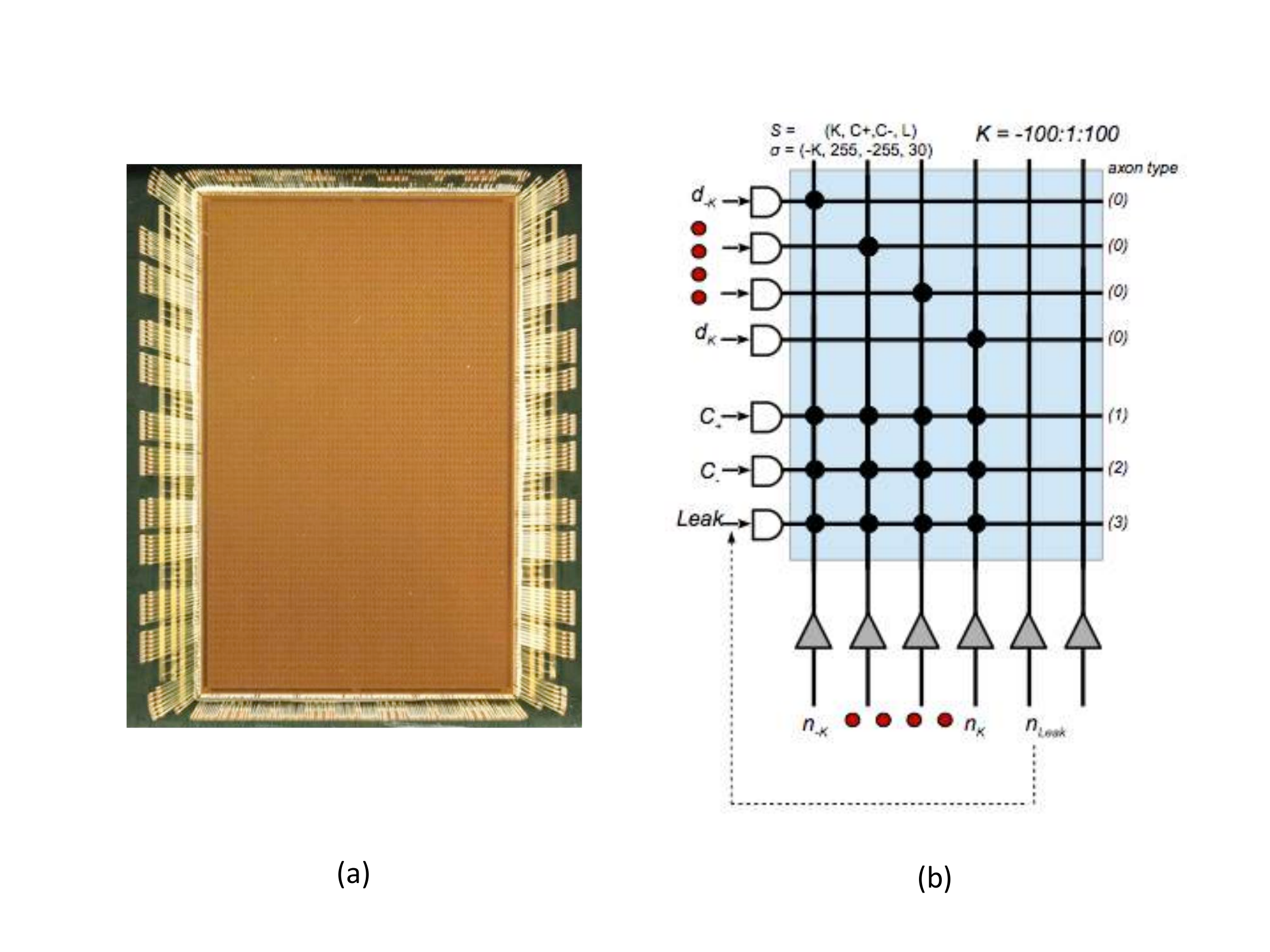}
  \caption{a) TrueNorth processor \ b) Sigmoid generation circuit with TrueNorth neurons and synapses}
\label{figure_sigmoid_char}
\end{figure}

\begin{figure}[!t]
  \centering
  \includegraphics[width=3.6in, height=1.5in]{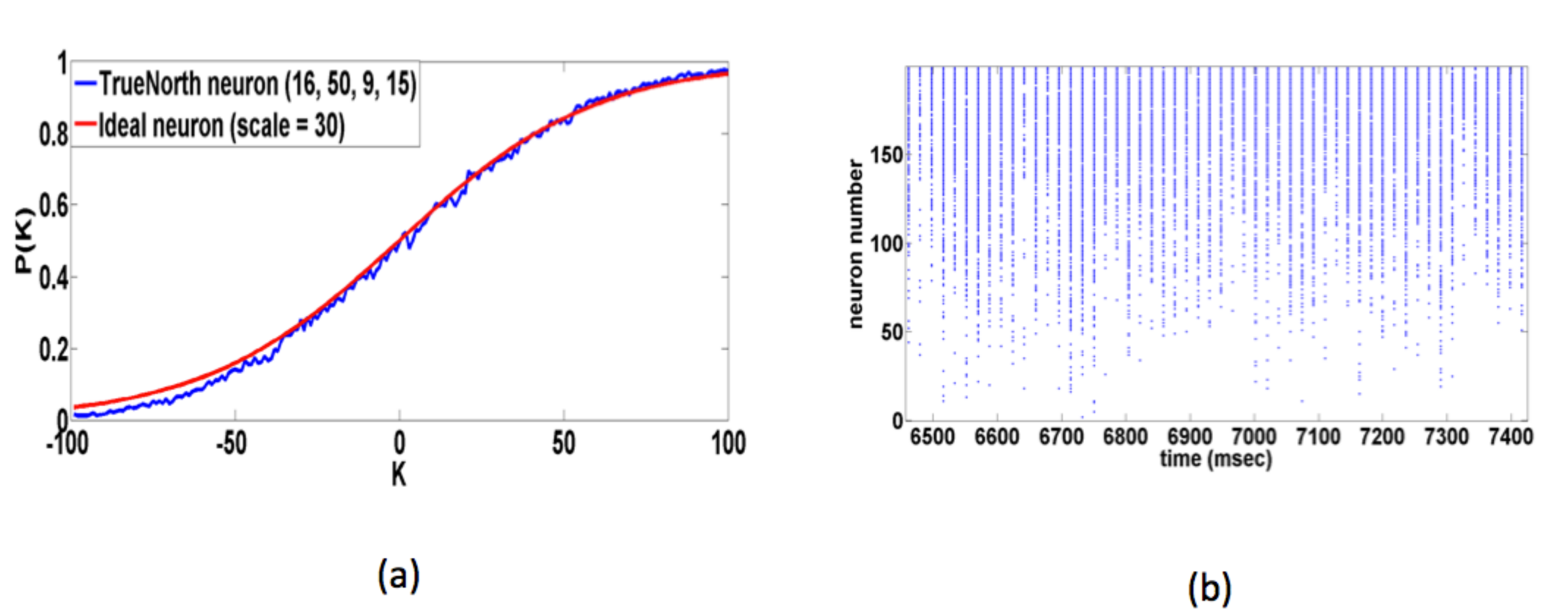}
  \caption{a)  Ideal sigmoid vs. TrueNorth realization \ b) Firing pattern showing how synaptic weight values ranging from -100 to 100 are used with the respective connected axons to activate data neurons 0 through 200 }
\label{figure_spike_raster}
\end{figure}

We implemented the digital sigmoid algorithm discussed here on TrueNorth as  shown in Fig. \ref {figure_sigmoid_char} using supported parameter values. Deterministic spike inputs are driven on axons $d_{-K}$ through $d_{K} (K$=$100) $ every Tw+2 (Tw=16) cycles and are integrated on the connected neurons $n_{-K}$ through $n_K$.  The synaptic weights are set between  -100 to 100 in steps of 1 for these connections (indicated by dots on the crossbar). Each such neuron is configured with a stochastic threshold $\alpha$ realized by $Vt$ = 50, $TM$ = 9 and leak value $\lambda$ set to 0. 
A single neuron $n_{leak}$ generates a stochastic leak of 0 or 1 for all the data neurons. This is multiplied by a factor of 15 by the weight on the connected axons and is the value of $leak$ used in the algorithm. The number of sampling intervals $Tw$ for a single application of the stimulus (spike input) is set to 16. Spiking probabilities obtained with TrueNorth neurons are compared versus the ideal sampler in Fig. \ref{figure_spike_raster}(a). A snapshot of the firing pattern of the characterized data neurons $n_{-K}$ through $n_{K}$ is shown in Fig. \ref{figure_spike_raster}(b).

\captionsetup{belowskip=-15pt,aboveskip=4pt}
\DeclareGraphicsExtensions{.pdf}

\DeclareGraphicsExtensions{.pdf}

\section {CONCLUSIONS}

We have demonstrated that approximate realizations of a Gibbs sampler on a digital neuromorphic substrate are feasible for classification with low system latency ($Tw = 1$). However, for generative tasks it may be necessary to use a form of the sampler that replicates the sigmoid more accurately via the use of a nonzero stochastic leak. Our proposed method of realization of the sigmoidal function with low-power, digital integrate-and-fire neurons is well suited for Gibbs sampling in RBMs and DBNs with parallel arrays of visible and hidden neurons in contrast to hardware implementations of sigmoids on standard Von Neumann computing platforms \cite{tommiska2003efficient, tisan2009digital}. Another advantage of our proposed implementation is that the required noise generation mechanism uses on-chip PRNG circuits which are easier to realize as compared to sampling from I\&F neurons with Gaussian noise \cite{neftci2013event}. Given these advantages such a sampler is ideally suited for a whole range of inference tasks in practical low-power, realtime applications which is the subject of ongoing investigation.

\section*{Acknowledgments}

The authors would like to thank the team members of the Brain-Inspired Computing group at IBM Almaden for supporting this project.



%



\bibliography{srinjoy}
\bibliographystyle{IEEEtran}

\end{document}